\documentclass[11pt]{article}
\usepackage{graphicx}
\usepackage{color}
\usepackage{amsmath}
\usepackage{amssymb}
\usepackage{amscd}
\usepackage{framed}
\usepackage{bbm}

\usepackage{fancyhdr,a4wide}
\usepackage{amsthm}

\newcommand{\R}{\mathbb{R}}
\newcommand{\inr}[1]{\left\langle #1 \right\rangle}

\newcommand{\E}{\mathbb{E}}

\newcommand{\eps}{\varepsilon}

\newtheorem{Theorem}{Theorem}[section]
\newtheorem{Lemma}[Theorem]{Lemma}

\newtheorem{Definition}[Theorem]{Definition}

\newtheorem{Corollary}[Theorem]{Corollary}
\newtheorem{Remark}[Theorem]{Remark}

\newtheorem{Assumption}{Assumption}[section]

\numberwithin{equation}{section}

\def \proof {\noindent {\bf Proof.}\ \ }

\def \endproof
{{\mbox{}\nolinebreak\hfill\rule{2mm}{2mm}\par\medbreak}}
\def\IND{\mathbbm{1}}

\def\IND{\mathbbm{1}}

\begin{document}
\title{Approximating the covariance ellipsoid}
\author{
Shahar Mendelson \thanks{Mathematical Sciences Institute, The Australian National University, and Department of Mathematics, Technion, I.I.T. Email: shahar.mendelson@anu.edu.au}}

\maketitle

\begin{abstract}
We explore ways in which the covariance ellipsoid ${\cal B}=\{v \in \R^d : \E \inr{X,v}^2 \leq 1\}$ of a centred random vector $X$ in $\R^d$ can be approximated by a simple set. The data one is given for constructing the approximating set consists of $X_1,...,X_N$ that are independent and distributed as $X$.

We present a general method that can be used to construct such approximations and implement it for two types of approximating sets. We first construct a (random) set ${\cal K}$ defined by a union of intersections of slabs $H_{z,\alpha}=\{v \in \R^d : |\inr{z,v}| \leq \alpha\}$ (and therefore ${\cal K}$ is actually the output of a neural network with two hidden layers). The slabs are generated using $X_1,...,X_N$, and under minimal assumptions on $X$ (e.g., $X$ can be heavy-tailed) it suffices that $N = c_1d \eta^{-4}\log(2/\eta)$ to ensure that $(1-\eta) {\cal K} \subset {\cal B} \subset (1+\eta){\cal K}$. In some cases (e.g., if $X$ is rotation invariant and has marginals that are well behaved in some weak sense), a smaller sample size suffices: $N = c_1d\eta^{-2}\log(2/\eta)$.

We then show that if the slabs are replaced by randomly generated ellipsoids defined using $X_1,...,X_N$, the same degree of approximation is true when $N \geq c_2d\eta^{-2}\log(2/\eta)$.

The construction we use is based on the small-ball method.
\end{abstract}

\section{Introduction}
Identifying the covariance of a centred random vector using random data is of central importance in high-dimensional statistics and has been studied extensively in recent years. The hope is that by using a relatively small sample $X_1,...,X_N$ of independent random vectors distributed as $X$, one can construct a good enough approximation of the covariance of $X$, and that such an approximation would be possible under minimal assumptions. The question is finding a `right way' of generating an approximation and then estimating the resulting tradeoff between the given sample size $N$, the degree of approximation and the probability with which that degree of approximation can be guaranteed.
\vskip0.4cm
The random vector $X$ endows an $L_2$ norm on $\R^d$ by setting for $v \in \R^d$,
$$
\|v\|_{L_2} \equiv \|\inr{X,v}\|_{L_2} = \left(\E (\inr{X,v})^2\right)^{1/2},
$$
and the unit ball of that norm is
$$
{\cal B} = \{v \in \R^d : \|v\|_{L_2} \leq 1\}=\{v \in \R^d : \inr{Tv,v}^{1/2} \leq 1\},
$$
where $T = \E (X \otimes X)$ is the covariance matrix of $X$. Throughout this note we assume without loss of generality that $T$ is invertible.

Given $X_1,...,X_N$ that are independent and distributed as $X$, a natural option is to consider the empirical covariance matrix
$\hat{T}=\frac{1}{N} \sum_{i=1}^N X_i \otimes X_i$ and approximate ${\cal B}$ by the random ellipsoid
$$
\hat{\cal B} = \left\{ v \in \R^d : \inr{\hat{T}v,v}^{1/2} \leq 1 \right\}.
$$
Note that even if one selects $\hat{\cal B}$ as the approximating set, there are various notions of approximation that one may consider. For example, by ensuring that the operator norm $\|\hat{T}-T\|_{2 \to 2} \leq \eta$, it follows that
$$
{\cal B} \subset \hat{\cal B} + \eta B_2^d \ \ \ \ {\rm and} \ \ \ \  \hat{\cal B} \subset {\cal B} + \eta B_2^d,
$$
where $B_2^d$ is the Euclidean unit ball and $A+B$ is the Minkowski sum $\{a+b : a \in A, \ b \in B\}$.

A different notion of approximation, which is the one that we focus on here, is equivalence between sets:
\begin{Definition} \label{def:approx}
The set ${\cal K} \subset \R^d$ is an $\eta$-approximation of ${\cal B}$ if
\begin{equation} \label{eq:geo-approx}
(1-\eta){\cal K} \subset {\cal B} \subset (1+\eta){\cal K}.
\end{equation}
\end{Definition}
For the choice of ${\cal K}= \hat{\cal B}$ an equivalent formulation of $\eta$-approximation is that
\begin{equation} \label{eq:quadratic-1}
\sup_{v \in {\cal B}} \left|\frac{1}{N}\sum_{i=1}^N \inr{X_i,v}^2 - \E \inr{X,v}^2 \right| \leq \eta.
\end{equation}
%%%%%%%%%%%%%%%%%%%%%%%%%%%%%%%%%%%%%%%%%%%%%%%%%%%%%%%%%%%%%5
Observe that if $T=\E (X \otimes X)$ then ${\cal B} = T^{-1/2}B_2^d$; hence, the random vector $Y=T^{-1/2}X$ is \emph{isotropic}: $\E (Y \otimes Y) = Id$, i.e, for every $v \in \R^d$, $\|\inr{Y,v}\|_{L_2} = \|v\|_2$. Moreover, denoting the Euclidean unit sphere by $S^{d-1}$, \eqref{eq:quadratic-1} becomes
\begin{equation} \label{eq:quadratic-2}
\sup_{v \in S^{d-1}} \left|\frac{1}{N}\sum_{i=1}^N \inr{Y_i,v}^2 - 1 \right| \leq \eta.
\end{equation}

The behaviour of \eqref{eq:quadratic-2}, the quadratic empirical process indexed by the unit sphere, is well understood (see e.g. \cite{MR2601042,MR3127875, MR3191978}). It characterizes the extremal singular values of the random matrix $N^{-1/2}\sum_{i=1}^N \inr{Y_i,\cdot}e_i$,  and is determined by two factors: the growth of moments of linear functionals $\inr{Y,v}$, and tail estimates on the Euclidean norm $\|Y\|_2$. The best known estimate on \eqref{eq:quadratic-2} in a heavy-tailed situation is due to Tikhomirov \cite{Tikh}:

\begin{Theorem} \label{thm:quad-est}
Let $Y$ be a centred, isotropic random vector in $\R^d$ and for $p>2$ set $L = \sup_{v \in S^{d-1}} \|\inr{Y,v}\|_{L_p}$. Let $Y_1,...,Y_N$ be independent, distributed according to $Y$. If  $\hat{T}=N^{-1}\sum_{i=1}^N Y_i \otimes Y_i$ then with probability at least $1-1/d$,
$$
C^{-1}\|Id-\hat{T}\|_{2 \to 2} \leq \frac{1}{N} \max_{1 \leq i \leq N} \|Y_i\|_2^2+\left(\frac{d}{N}\right)^{1-2/p} \log^4\left(\frac{eN}{d}\right)+\left(\frac{d}{N}\right)^{1-2/\min\{4,p\}},
$$
for a constant $C$ that depends only on $L$ and $p$.
\end{Theorem}

If one believes that Theorem \ref{thm:quad-est} is reasonably sharp, it casts a shadow on the choice of $\hat{\cal B}$ as an $\eta$-approximation of ${\cal B}$ in the sense of Definition \ref{def:approx}. Indeed, when $X$ is heavy-tailed it is likely that some of the vectors $Y_i=T^{-1/2}X_i$ will have large Euclidean norms. In Section \ref{sec:example} we will give a concrete example of an isotropic random vector that satisfies an $L_4-L_2$ norm equivalence, but still $\hat{\cal B}$  is very different from ${\cal B}$ with a non-negligible probability.

Of course, while $\hat{\cal B}$ is the natural choice for a data-dependent approximation of ${\cal B}$, it is certainly not the only choice. For one, there is no reason to restrict the approximating set to an ellipsoid, though it is not clear offhand how one may generate other approximating sets given the limited data at one's disposal.

The method we present does just that. Its starting point is identifying a random property that is satisfied only by points in a set that is `close enough' to ${\cal B}$. To give an example of what we mean by a random property, assume, for example, that $X$ is the standard gaussian vector in $\R^d$. Then ${\cal B}=B_2^d$, and for each $v \in \R^d$, $\inr{X,v}$ is a centred gaussian random variable whose variance is $\|v\|_2^2$. Thus, using the values $\inr{X_1,v},...,\inr{X_N,v}$ one may identify $\|v\|_2$ rather accurately and in particular pin-point the Euclidean sphere on which $v$ is located. The difficulty lies in the fact that the accurate estimate has to hold uniformly for every $v \in \R^d$, and how that can be achieved is not obvious. Our method leads to such uniform estimates, and as examples we obtain approximation results using two different types of sets.

\vskip0.4cm
The first example we consider has to do with approximations generated by slabs. For $z \in \R^d$ and $\alpha>0$ set $H_{z,\alpha} = \{v \in \R^d : |\inr{z,v}| \leq \alpha\}$. Given $z_1,...,z_n \in \R^d$ and $\alpha_1,...,\alpha_n>0$,  define
$$
{\cal K} = \{v \in \R^d : v \in H_{z_j,\alpha_j} \ {\rm for \ at \ least \ } \beta n \ {\rm indices} \}.
$$
In other words, ${\cal K}$ is a union of all the intersections of $\beta n$ slabs out of $(H_{z_i,\alpha_i})_{i=1}^n$. Note that ${\cal K}$ need not be a convex set though it is star-shaped around $0$: if $v \in {\cal K}$ then for any $0 \leq \theta \leq 1$, $\theta v \in {\cal K}$.

This type of approximation has been studied in \cite{MR1761898}, where the authors attempted to approximate the characteristic function of the Euclidean unit ball in $\R^d$ by the characteristic function of a simple set. It was well known that approximating the Euclidean unit ball by a polytope required the polytope to have at least $\exp(cd)$ faces (see, e.g., \cite{MR608101,MR670396} for accurate statements), and the alternative studied in \cite{MR1761898} was to approximate $\IND_{B_2^d}$ by the output of a neural network with two hidden layers; that is, by a characteristic function of a set of the form
\begin{equation} \label{eq:K1}
\left\{ v \in \R^d : \sum_{i=1}^n \gamma_i \IND_{\{\inr{z_i,v} \geq \alpha_i\}} \geq k \right\}.
\end{equation}
It was shown in \cite{MR1761898} that one may construct such a set ${\cal K}_1$ using $n=cd^2/\eta^2$ points $z_i$, and for the right choice of $\alpha_i$ and $\gamma_i$ one has
$$
(1-\eta)B_2^d \subset {\cal K}_1 \subset (1+\eta)B_2^d.
$$
Unfortunately, although it is possible to derive a similar approximation for a general ellipsoid, that construction requires information on the ellipsoid's principal axes, making it unhelpful for covariance approximation.

In \cite{MR2204286} the authors considered similar approximating sets (which they called `zig-zag bodies'), but their approach for choosing the points $z_i$ and thresholds $\alpha_i$ was more promising from our perspective; moreover, it also led to a better estimate on the required number of slabs.

\begin{Theorem} \label{thm:zig-zag} \cite{MR2204286}
There exist absolute constants $c_1$ and $c_2$ for which the following holds. Let $Z$ be distributed according to the uniform measure on $S^{d-1}$
and let $Z_1,...,Z_N$ be independent, distributed as $Z$. Set
\begin{equation} \label{eq:zig-zag-body}
{\cal K}_2 = \left\{v \in \R^d : |\inr{v,Z_i}| \leq \alpha_d \ \ {\rm for \ at \ least \ } N/2 \ {\rm indices}\right\},
\end{equation}
where $\alpha_d$ is the median of $|\inr{Z,v}|$ for $v \in S^{d-1}$. If $0<\eta<1$ and $N = c_1 d\eta^{-2}\log(2/\eta)$ then with probability at least $1-2\exp(-c_2d)$,
$$
(1-\eta)B_2^d \subset {\cal K}_2 \subset (1+\eta)B_2^d.
$$
\end{Theorem}
In other words, the Euclidean ball (which, up to a normalization factor of $c_d \sqrt{d}$, $\lim_{d \to \infty} c_d=1$, is the covariance unit ball endowed by $Z$) can be approximated by the union of intersections generated by $c(\eta)d$ slabs, and this approximation holds with very high (exponential) probability.

\begin{Remark}
Note that ${\cal K}_2$ belongs to the family of sets \eqref{eq:K1}. Indeed, this is evident because
$$
{\cal K}_2 = \left\{v \in \R^d : \sum_{i=1}^N \IND_{\{|\inr{v,Z_i}| \leq \alpha_d\}} \geq \frac{N}{2} \right\},
$$
and for $\alpha>0$, $\IND_{\{|\inr{v,z}| \leq \alpha\}} = \IND_{\{\inr{v,z} \geq - \alpha\}}-\IND_{\{\inr{v,z} \geq \alpha\}}$.
\end{Remark}

\vskip0.4cm
The proof of Theorem \ref{thm:zig-zag} relies heavily on the fact that $Z_1,...,Z_N$ are distributed according to the uniform measure on the sphere. However, it still opens the door to a possible way of addressing the problem at hand: one may try to select ${\cal K}$ randomly, in a similar way to \eqref{eq:zig-zag-body}.

We will show that indeed Theorem \ref{thm:zig-zag} can be extended---with some necessary modifications---to an almost arbitrary centered random vector. The proof is based on a random property that allows one to check accurately whether $v \in \R^d$ actually belongs to ${\cal B}$ or not. As we explain in what follows, that property is reflected by the `frequency' with which the $X_i$'s belong to an appropriate slab defined by $v$ (see Section \ref{sec:small-ball} for details).

\vskip0.4cm

To formulate our main results we need to introduce some additional notation.  Throughout, absolute constants are denoted by $c,c_0,c_1,...$; their values may change from line to line. $a \lesssim b$ means that there is an absolute constant $c$ such that $a \leq cb$, and $a \sim b$ implies that $ca \leq b \leq Ca$ for absolute constants $c$ and $C$. Finally, $a \sim_L b$ denotes that $ca \leq b \leq Ca$ for constants $c$ and $C$ that depend only on $L$.

Given integers $m$ and $n$ set $N=nm$. Let $\{X_{i,j} : 1 \leq i \leq m, 1 \leq j \leq n\}$ be $N$ independent copies of $X$ and for $1 \leq j \leq n$ put
$$
Z_j = \frac{1}{\sqrt{m}} \sum_{i=1}^m X_{i,j}.
$$
Also, denote by $g$ the standard gaussian random variable and set $\alpha$ to be the median of $|g|$. For $\eta>0$ define the random set
$$
{\cal K}_\eta = \left\{v \in \R^d : |\inr{Z_j,v}| \leq \alpha + \eta \ {\rm for \ at \ least \ } \left(\frac{1}{2}-\eta\right)n \ {\rm indices} \ j \right\}.
$$
\begin{Theorem} \label{thm:main-intro}
Let $0<\eta<1/10$ and $L \geq 1$. Assume that for every $v \in \R^d$, $\|\inr{X,v}\|_{L_q} \leq L\|\inr{X,v}\|_{L_2}$ for some $q>2$, set $m \geq c_0(\eta,L)$ and let $n \geq c_1(\eta)d$.

Then, with probability at least $1-2\exp(-c_2 \eta^2 n)$,
$$
{\cal B} \subset {\cal K}_\eta \subset (1+c_3\eta){\cal B},
$$
for absolute constants $c_2$ and $c_3$.

Moreover, if $q \geq 3$ one may take
$$
c_0 \sim_L \eta^{-2} \ \ {\rm and} \ \ c_1 \sim \eta^{-2}\log(2/\eta),
$$
implying that $N =c(L)d \eta^{-4}\log(2/\eta)$ points suffice.
\end{Theorem}
As it happens, the superfluous factor of $\log(2/\eta)$ can be removed from Theorem \ref{thm:main-intro} if one employs a different method of proof. However, the required argument is rather specific and holds only for approximation by slabs as in Theorem \ref{thm:main-intro}. Because the main point of this note is to advocate our method of constructing approximations, we chose to present the general argument and only outline the alternative proof of Theorem \ref{thm:main-intro} (see Section \ref{sec:improve}).

\begin{Remark}
As we explain in what follows, if $X$ is a `nice' random vector (and among these `nice' random vectors are the standard gaussian vector or the vector distributed uniformly on the Euclidean unit sphere) then one may take $m=1$ and $n \sim d \eta^{-2}\log(2/\eta)$ (or $n \sim d \eta^{-2}$ using the alternative proof). In particular, Theorem \ref{thm:main-intro} improves Theorem \ref{thm:zig-zag}.
\end{Remark}

In the other example we present we construct a more complex approximating set: it is the union of intersections of ellipsoids rather than the union of intersections of slabs. On the other hand, the required sample size is smaller and all that one needs is the following weak assumption on $X$:
\begin{Assumption} \label{ass:ellipsoids}
Assume that for every $\eta>0$ there is some $m=m_0(\eta)$ for which the following holds: if $\|v\|_{L_2}=1$ then
$$
Pr\left( \left| \frac{1}{m}\sum_{i=1}^m \inr{X_i,v}^2 - 1\right| \geq \frac{\eta}{10} \right) \leq 0.01.
$$
\end{Assumption}

To see that Assumption \ref{ass:ellipsoids} is rather minimal, note that under an $L_4-L_2$ norm equivalence (i.e., that for every $v \in \R^d$, $\|\inr{X,v}\|_{L_4} \leq L\|\inr{X,v}\|_{L_2}$), one has $m_0(\eta) \leq c(L)/\eta^2$. Naturally, nontrivial estimates on $m_0(\eta)$ are possible in more general situations than an $L_4-L_2$ norm equivalence.

The `ellipsoid approximation' estimate is as follows:
\begin{Theorem} \label{thm:ellipsoids}
There exist absolute constants $c_0,c_1$ and $c_2$ for which the following holds. For $0<\eta<1/4$ let $m=m_0(\eta)$ and $n \geq c_0\max\{d\log(m/\eta),m\}$. Put $N = nm$ and set $(X_{i,j}), \ 1 \leq i \leq m, \ 1 \leq j \leq n$ to be independent, distributed according to $X$. If

$$
{\cal D}_\eta = \left\{v \in \R^d : \frac{1}{m}\sum_{i=1}^m \inr{X_{i,j},v}^2 \leq 1+\eta \ {\rm for \ at \ least \ } 0.9n \ {\rm indices} \ j \right\},
$$
then with probability at least $1-2\exp(-c_1 n/m)$,
$$
{\cal B} \subset {\cal D}_\eta \subset (1+c_2\eta){\cal B}.
$$
\end{Theorem}
To put the outcome of Theorem \ref{thm:ellipsoids} in some perspective, under an $L_4-L_2$ norm equivalence one has that $m_0(\eta) \leq c(L)/\eta^2$, implying that $n=c^\prime \max\{d \log(L/\eta),\eta^{-2}\}$ suffices, and the resulting required sample size of $N  \sim d \eta^{-2} \log(2/\eta)$ is better than the outcome of Theorem \ref{thm:main-intro} by a factor of $1/\eta^2$ as long as $\eta \geq 1/d^{1/2}$.

\vskip0.4cm
In the next section we describe the general method and explain how it is used in the proofs of Theorem \ref{thm:main-intro} and Theorem \ref{thm:ellipsoids}. The argument is actually a variant of the small-ball method introduced in \cite{MenACM}. The proofs of Theorem \ref{thm:main-intro} and Theorem \ref{thm:ellipsoids} are presented in Section \ref{sec:proofs}.

\section{The small-ball method} \label{sec:small-ball}
Let us begin by describing the argument used in the proof of Theorem \ref{thm:zig-zag}. It is based on three crucial observations:
\begin{description}
\item{$\bullet$} \emph{All the points on a centred sphere behave in the same way:} By rotation invariance, if $Z$ is distributed according to the uniform measure on $S^{d-1}$ then all the random variables $\inr{Z,v/\|v\|_2}$ have the same distribution;  therefore $|\inr{Z,v/\|v\|_2}|$ all have the same quantiles, and in particular, the same median.

\item{$\bullet$} \emph{Quantiles can be used to `separate' between different spheres:} If $\|u\|_2 \not = \|v\|_2$, that fact is reflected in a difference between $Pr(|\inr{Z,v}| \leq \alpha)$ and $Pr(|\inr{Z,u}| \leq \alpha)$.

\item{$\bullet$} \emph{Separation is visible through sampling:} For every $v \in \R^d$, the sum of independent indicators
$$
\frac{1}{N} \sum_{i=1}^N \IND_{\{|\inr{Z,v}| \leq \alpha\}}
$$
exhibits sharp concentration around $Pr(|\inr{Z,v}| \leq \alpha)$.
\end{description}

It follows that for every $v \in S^{d-1}$, the median $\alpha_d$ of $|\inr{Z,v}|$ is the same (and happens to be $c_d/\sqrt{d}$ with $\lim_{d \to \infty} c_d =1$). Moreover, given $Z_1,...,Z_N$ that are independent and distributed according to $Z$, $|\{j: |\inr{Z_j,v}| \leq \alpha_d\}|$ is highly concentrated around $N/2$.

The heart of the proof is to show that a similar bound is true \emph{uniformly} on $S^{d-1}$; that is, with high probability,
\begin{equation} \label{eq:zig-zag-concentration}
\sup_{v \in S^{d-1}} \left| \left| \left\{j : |\inr{Z_j,v}| \leq \alpha_d \right\} \right| -\frac{N}{2}  \right|
\end{equation}
is small provided that $N$ is large enough.

To establish \eqref{eq:zig-zag-concentration}, note that the high probability estimate that holds for every individual $v$ allows one to obtain uniform control on a fine enough net in $S^{d-1}$. And, if $\pi u$ denotes the best approximation to $u$ in the net, $|\inr{Z_j,u}|$ cannot be different from $|\inr{Z_j,u-\pi u}|$ by much; indeed, $|\inr{Z,u-\pi u}| \leq \|Z\|_2 \|u-\pi u\|_2 = \|u-\pi u\|_2$ because $Z$ is supported on $S^{d-1}$.

Once \eqref{eq:zig-zag-concentration} is established, the outcome of Theorem \ref{thm:zig-zag} follows immediately: the set
$$
{\cal K}_2=\left\{v \in \R^d : |\inr{v,z_i}| \leq \alpha_d \ \ {\rm for \ at \ least \ } N/2 \ {\rm indices}\right\}
$$
contains $(1-\eta)S^{d-1}$, but does not contain any point on $(1+\eta)S^{d-1}$. Therefore, since ${\cal K}_2$ is star-shaped around $0$, $(1-\eta)B_2^d \subset {\cal K}_2 \subset (1+\eta)B_2^d$.

\vskip0.4cm

It is clear that when dealing with a general random vector, most of the features used in the proof of Theorem \ref{thm:zig-zag} are simply not true: quantiles $Pr(|\inr{X,v}| \leq \alpha)$ may change on the $L_2$ unit sphere
$$
{\cal S}=\{v \in \R^d : \E|\inr{X,v}|^2=1\};
$$
they need not `separate' between two $L_2$ spheres; and `oscillations' $|\inr{X_i,u-\pi u}|$ can be large, especially when $X$ is heavy-tailed rather than being bounded like in Theorem \ref{thm:zig-zag}.

\vskip0.4cm

The analysis required for addressing these difficulties is based on the \emph{small-ball method}, which was introduced in \cite{MenACM} to deal precisely with this sort of problem: obtaining high probability, uniform estimates in heavy-tailed situations. The path we take follows the main ideas of the method:
\begin{description}
\item{$(a)$} Identify a property ${\cal P}$ that allows one to check whether a fixed $v \in \R^d$ belongs to ${\cal B}$ or not - using only the probability with which the property holds. Moreover, ${\cal P}$ should be defined using only on relatively small number of the independent copies of $X$ at one's disposal.

    For example, one may consider the functionals
$$
 \frac{1}{\sqrt{m}} \sum_{i=1}^m \inr{X_i,v}  \ \ \ \ \ {\rm and} \ \ \ \ \ \frac{1}{m}\sum_{i=1}^m \inr{X_i,v}^2
$$
where $m$ is relatively small. The former is close to a centred gaussian variable whose variance is $\E \inr{X,v}^2=\|v\|_{L_2}^2$ while the latter concentrates around $\|v\|_{L_2}^2$. Therefore, if the goal is to check whether $\|v\|_{L_2} \leq 1$ one may define
\begin{equation} \label{eq:property-P}
{\cal P}_1 =  \left\{\left|\frac{1}{\sqrt{m}} \sum_{i=1}^m \inr{X_i,v} \right| \leq \alpha+\eta \right\} \ \ {\rm and} \ \
{\cal P}_2 = \left\{\frac{1}{m}\sum_{i=1}^m \inr{X_i,v}^2  \leq 1+ \frac{\eta}{10}\right\}
\end{equation}
respectively, where $\alpha$ appearing in ${\cal P}_1$ is the median of $|g|$, the absolute value of a standard gaussian, and $\eta$ is small.

In both cases the probability of the events in question are determined by $\|v\|_{L_2}$: the probability of ${\cal P}_1$ will be very close to $1/2$ if and only if $\|v\|_{L_2} =1$, whereas ${\cal P}_2$ holds with probability that is close to $1$ if $\|v\|_{L_2} \leq 1+\eta$ and with probability that is close to $0$ in $\|v\|_{L_2}$ is much larger.
\end{description}
 In general, the idea in $(a)$ is that the identity of $\|v\|_{L_2}$ is reflected by the probability with which ${\cal P}$ hold. The next step is to `detect' that probability with very high confidence.

\begin{description}
\item{$(b)$}  Split $\{1,...,N\}$ to $n$ coordinate blocks $I_j$, each one of cardinality $m$ and set $W_j(v)=\IND_{\{v \ {\rm statisfies \ } {\cal P}\}}(X_i, \ i \in I_j)$. It is evident that $W(v)=n^{-1}\sum_{j=1}^n W_j(v)$ concentrates around its mean, i.e., the probability with which ${\cal P}$ holds. Therefore, the cardinality $|\{j : W_j(v)=1\}|$ leads to a very good estimate of that probability, and in particular of $\|v\|_{L_2}$. Moreover, the resulting estimate is valid with confidence that is exponential in $n=N/m$, say $1-2\exp(-cn)$.

\item{$(c)$}  Use $(b)$ to define the random approximating set ${\cal K}$: \emph{$v$ belongs to the set if $W_j(v)=1$ for the `right number' of indices $j$}.

\end{description}
    Now one needs to verify that the resulting set ${\cal K}$ is truly close to ${\cal B}$. If ${\cal K}$  happens to be star-shaped around $0$, it suffices to ensure that ${\cal S} \subset {\cal K}$, and at the same time that $\{ v : \|v\|_{L_2} = 1+\eta\} \subset {\cal K}^c$.  As a result, one has to obtain a uniform estimate on the cardinality $|\{j : W_j(v)=1\}|$ for $v$'s that belong to the two centred $L_2$ spheres: the unit one, and the one of radius $1+\eta$:
\begin{description}
\item{$(d)$} The high probability estimate with which $(b)$ holds allows one to control a large collection of $v$'s uniformly. The obvious choice of such a set $V$ is an appropriate $L_2$-net in the sphere in question. This leads to an estimate that holds with high confidence but only for points in $V$ rather than for the entire sphere.
\item{$(e)$} Finally, to pass from $V$ to the entire sphere one must control the oscillations:  show that if $u$ is `close' to $v$, then the number of indices $j$ on which $W_j(u)=1$ is very close to the number of indices on which $W_j(v)=1$.
\end{description}
Clearly, the key step is $(e)$: obtaining the required uniform control on random oscillations, a task that is nontrivial in heavy-tailed situations.

\vskip0.4cm
As this description indicates, the method is rather general and can be employed for a wide variety of choices of ${\cal P}$. One may consider other alternatives beyond the two examples we present in what follows, and those would result in different approximating sets. The crucial point is that as long as ${\cal P}$ is well chosen, those sets would all be good approximations of the covariance ellipsoid.

\section{Proofs} \label{sec:proofs}
Before we present the proofs of Theorem \ref{thm:main-intro} and Theorem \ref{thm:ellipsoids} we need the following standard observation:

\begin{Lemma} \label{lemma:Bernoulli}
Let $X$ be a centred random vector in $\R^d$ and let $X_1,...,X_k$ be independent copies of $X$. Then
$$
\E \sup_{v \in {\cal B}} \left|\sum_{i=1}^k \eps_i \inr{X_i,v} \right| \leq \sqrt{k} \sqrt{d},
$$
where $(\eps_i)_{i=1}^k$ are independent, symmetric, $\{-1,1\}$-valued random variables that are independent of $X_1,...,X_k$.
\end{Lemma}

\proof Let $T = \E (X \otimes X)$, and recall that ${\cal B} = T^{-1/2}B_2^d$ and that $T^{-1/2}X$ is isotropic. Note that for an isotropic vector $Y$,
$$
\E \|Y\|_2^2 = \E \sum_{i=1}^d \inr{Y,e_i}^2 = d.
$$
Therefore,
\begin{align*}
& \E \sup_{v \in {\cal B}} \left|\sum_{i=1}^k \eps_i \inr{X_i,v} \right| = \E \sup_{w \in B_2^d} \left|\sum_{i=1}^k \eps_i \inr{X_i,T^{-1/2}v} \right|
\\
& =  \E \sup_{w \in B_2^d} \left|\sum_{i=1}^k \eps_i \inr{T^{-1/2}X_i,v} \right| \leq \E_X \left(\sum_{i=1}^k \|T^{-1/2}X_i\|_2^2 \right)^{1/2},
\end{align*}
and the claim follows from Jensen's inequality and the fact that $T^{-1/2}X$ is isotropic.
\endproof

\subsection{Approximation by slabs} \label{sec:approx-slab}
Recall that ${\cal S} \subset \R^d$ is the $L_2$ unit sphere; that is, ${\cal S}=\{v \in \R^d : \|\inr{X,v}\|_{L_2} =1 \}$.

As a starting point, let $Z$ be a random vector that has the same covariance as $X$, and therefore endows the same $L_2$ structure on $\R^d$---in particular, $Z$ endows the same unit ball ${\cal B}$ and unit sphere ${\cal S}$.  Assume that there are $\alpha>0$, $0<\beta<1$, $\eta < \beta/4$, $\eps_0 < \alpha/2$ and $\gamma>6\eta/\alpha$ such that for every $v \in {\cal S}$ and every $\eps_0<\eps<\alpha/2$,
\begin{description}
\item{$(1)$} $|Pr(|\inr{Z,v}| \leq \alpha) -\beta| \leq \eta$, and
\item{$(2)$} $Pr(|\inr{Z,v}| \in [\alpha-\eps,\alpha]) \geq \gamma \eps$.
\end{description}

To explain this condition, one should think of $\eta$ as a small number (measuring the wanted degree of approximation), and that $\alpha$ and $\beta$ are just constants; thus, Condition $(1)$ means that the function $\phi(v)=Pr(|\inr{Z,v}| \leq \alpha)$ is roughly a constant on the sphere ${\cal S}$. Condition $(2)$ means that the (marginal) mass of a small interval that ends at $\alpha$ is nontrivial; in other words, there is a noticeable difference between $Pr(|\inr{Z,v}| \leq \alpha)$ and  $Pr(|\inr{Z,v}| \leq \alpha-\eps)$ for every $v \in {\cal S}$; the lower bound on $\gamma$ is there to ensure that the difference between the two is indeed noticeable.

\vskip0.4cm

Note that $G$, the standard gaussian vector in $\R^d$, satisfies $(1)$ and $(2)$: ${\cal S}=S^{d-1}$; for every $v \in S^{d-1}$, $\inr{G,v}$ is distributed as a standard gaussian variable; and one may set $1/10 \leq \alpha \leq 10$, $\beta=Pr(|g| \leq \alpha)$, $\gamma$ that is an absolute constant and $\eps_0=0$. A similar argument shows that the uniform measure on $S^{d-1}$ also satisfies $(1)$ and $(2)$ for the right choice of constants.

As we explain in what follows, in general situations our choice of $Z$ will only have approximately gaussian one-dimensional marginals, and that would suffice to ensure that both $(1)$ and $(2)$ hold for $\alpha,\beta$ and $\gamma$ that are absolute constants.

\vskip0.4cm
The main component in the proof of Theorem \ref{thm:main-intro} is the next fact:

\begin{Theorem} \label{thm:main-1}
There exist constants $c_0,c_1,c_2$ that depend only on $\alpha,\beta$ and $\gamma$ for which the following holds. Let $Z$ satisfy $(1)$ and $(2)$ for some $\eps_0 \leq (3/\gamma)\eta$. Let $Z_1,...,Z_n$ be independent copies of $Z$ and set
$$
{\cal K} = \{v \in \R^d : |\inr{Z_j,v}| \leq \alpha + \eta \ {\rm for \ at \ least \ } (\beta-\eta)n \ {\rm indices} \ j\}.
$$
If $n \geq c_1 d\eta^{-2}\log(2/\eta)$ then with probability at least $1-2\exp(-c_2 n \eta^2)$,
$$
{\cal B} \subset {\cal K} \subset (1+c_3\eta){\cal B}
$$
\end{Theorem}

\proof We follow the path outlined in Section \ref{sec:small-ball}. Thanks to $(1)$ and $(2)$ we have the wanted property using a single copy of $Z$. Indeed, as a preliminary step observe that $\{|\inr{Z,v}|\leq \alpha\}$ holds with probability that does not change much on ${\cal S}$. At the same time, by the lower bound on $\gamma$, $\alpha/2 \leq \alpha-(3/\gamma) \eta < \alpha$, and fix $1<\rho \leq 2$ such that $\alpha/\rho = \alpha-(3/\gamma) \eta$. Since $\eps_0 \leq (3/\gamma)\eta$ it follows that $(2)$ holds for $\eps=(3/\gamma)\eta$ and one has
$$
Pr(|\inr{Z,v}| \leq \alpha/\rho) \leq \beta -  3\eta.
$$
Thus, there is a noticeable difference between $Pr(|\inr{Z,v}| \leq \alpha)$ and $Pr(|\inr{Z,\rho v}| \leq \alpha)$.

By Bernstein's inequality, it follows that with probability at least $1-2\exp(-c_0(\beta)n\eta^2)$,
$$
\left|\frac{1}{n}\sum_{j=1}^n \IND_{\{|\inr{Z_j,v}| \leq \alpha\}} - Pr(|\inr{Z,v}| \leq \alpha)\right| \leq \eta/2;
$$
Therefore, on that event,
\begin{equation} \label{eq:main-single-cond-1}
|\{j : |\inr{Z_j,v}| \leq \alpha\}| \geq n(\beta-\eta/2).
\end{equation}
Applying Bernstein's inequality again, with probability at least $1-2\exp(-c_0(\beta)\eta^2n)$,
\begin{equation} \label{eq:main-single-cond-2}
\left|\left\{j: |\inr{Z_j,v}| \leq \frac{\alpha}{\rho}\right\}\right| \leq \left(\beta - 2\eta\right)n.
\end{equation}

The heart of the proof is to show that slightly modified versions of \eqref{eq:main-single-cond-1} and \eqref{eq:main-single-cond-2} hold uniformly on ${\cal S}$; that is,  with high probability, for every $v \in {\cal S}$,
\begin{equation} \label{eq:main-single-cond-1-m}
|\{j : |\inr{Z_j,v}| \leq \alpha+\eta\}| \geq n(\beta-\eta),
\end{equation}
and
\begin{equation} \label{eq:main-single-cond-2-m}
\left|\left\{j: |\inr{Z_j,v}| \leq \frac{\alpha+\eta}{\rho}\right\}\right| < n(\beta-\eta).
\end{equation}

Let $c_1=c_0/2$ and let $V \subset {\cal S}$ be an maximal $r$-separated subset of ${\cal S}$ with respect to the $L_2$ norm and of cardinality at most $\exp(c_1\eta^2 n)$. There is an event ${\cal A}_1$ of probability at least $1-4\exp(-c_1 \eta^2 n)$ on which \eqref{eq:main-single-cond-1} and \eqref{eq:main-single-cond-2} hold for every $v \in V$. Also, because ${\cal B}$ is a convex, centrally-symmetric subset of $\R^d$,  a standard volumetric estimate shows that
\begin{equation} \label{eq:r}
r \leq 5\exp(-c_1\eta^2 n/d).
\end{equation}

For every $u \in {\cal S}$ let $\pi u$ be the nearest in $V$ to $u$ with respect to the $L_2$ norm. Set
$$
W = \sup_{u \in {\cal S}} \sum_{j=1}^n \IND_{\{|\inr{Z_j,u-\pi u}| \geq t\}}
$$
for $t=\eta/2$ (which is smaller than $\eta/\rho$).  Our aim is to ensure that with high probability $W \leq n\eta/2$, and to that end we first estimate $\E W$. Observe that
$$
W \leq \sup_{u \in {\cal S}} \frac{1}{t}\sum_{j=1}^n |\inr{Z_j,u-\pi u}|;
$$
by the Gin\'{e}-Zinn symmetrization theorem \cite{MR757767} followed by the contraction inequality for Bernoulli processes \cite{LeTa91},
\begin{align*}
\E W \leq & \frac{2}{t} \left(\E \sup_{u \in {\cal S}} \left|\sum_{j=1}^n \eps_j \inr{Z_j,u-\pi u} \right| + n \sup_{u \in {\cal S}} \E |\inr{Z,u-\pi u}|\right)
\\
\leq & \frac{2r}{t} \left(\E \sup_{u \in {\cal S}} \sum_{j=1}^n \eps_j \inr{Z_j,u} + n\right),
\end{align*}
where we have used the fact that $\|u-\pi u\|_{L_1} \leq  \|u-\pi u\|_{L_2} \leq r$. Moreover, by Lemma \ref{lemma:Bernoulli}, $\E \sup_{u \in {\cal S}} \sum_{j=1}^n \eps_j \inr{Z_j,u} \leq \sqrt{n}\sqrt{d}$, implying that if $n \geq d$ then
$$
\E W \leq \frac{c_2n}{t} \exp(-c_1 \eta^2n/d)= \frac{2c_2 n}{\eta} \exp(-c_1 \eta^2n/d),
$$
thanks to the estimate on $r$ from \eqref{eq:r} and by the choice of $t$.

Now, by the bounded differences inequality (see, e.g., \cite{BoLuMa13}), we have that for every $x>0$, $Pr(W \geq \E W +x) \leq \exp(-c_3x^2/n)$. Setting $x=n\eta/4$, there is an event ${\cal A}_2$ of probability at least $1-2\exp(-c_4 \eta^2n)$ on which
$$
W \leq n \left(\frac{2c_2}{\eta} \exp(-c_1\eta^2 n/d) + \frac{\eta}{4}\right) \leq \frac{\eta}{2}n,
$$
where the last inequality holds if we set
$$
n \gtrsim \frac{d}{\eta^2}\log\left(\frac{2}{\eta}\right).
$$

Combining the two estimates, on the event ${\cal A}_1 \cap {\cal A}_2$ one has that for any $u \in {\cal S}$ both \eqref{eq:main-single-cond-1-m} and \eqref{eq:main-single-cond-2-m} hold. Indeed, for every $u \in {\cal S}$ we have
\begin{description}
\item{$\bullet$} $|\inr{Z_j,\pi u}| \leq \alpha$ for at least $n(\beta-\eta/2)$ indices $j$; and
\item{$\bullet$} $|\inr{Z_j,u-\pi u}| \geq \eta$ for at most $\eta/2$ indices $j$.
\end{description}
Therefore, there is a set of indices of cardinality at least $n(\beta-\eta)$ such that both $|\inr{Z_j,\pi u}| \leq \alpha$ and $|\inr{Z_j,u-\pi u}| \leq \eta$, and for those indices,
$$
|\inr{Z_j,u}| \leq |\inr{Z_j,\pi u}|+|\inr{Z_j,u-\pi u}| \leq \alpha+\eta,
$$
verifying \eqref{eq:main-single-cond-1-m}. A similar argument may be used to confirm \eqref{eq:main-single-cond-2-m}.

Setting
$$
{\cal K} = \{v \in \R^d : |\{i : |\inr{Z_j,v}| \leq \alpha+\eta\}| \geq  (\beta-\eta)n\},
$$
it follows from \eqref{eq:main-single-cond-1-m} that ${\cal S} \subset {\cal K}$; and, since ${\cal K}$ is star-shaped around $0$, ${\cal B} \subset {\cal K}$ as well.

On the other hand, recalling that $\eta \leq \alpha \gamma/6$ then
$$
\rho = 1+\frac{3\eta}{\alpha\gamma -3\eta} \leq 1+c_5\eta,
$$
where $c_5 \sim 1/\alpha \gamma$. Thus, if $\|u\|_{L_2} =\rho > 1$, then
$$
\{j : |\inr{Z_j,u}| \leq \alpha+\eta\} = \left\{j:  |\inr{Z_j,v}| \leq \frac{\alpha+\eta}{\rho} \right\}
$$
for some $v \in {\cal S}$. Hence, by \eqref{eq:main-single-cond-2-m},
$$
|\{j : |\inr{Z_j,u}| \leq \alpha+\eta\}| < (\beta-\eta)n,
$$
and $u \not \in {\cal K}$. It follows that $\{v : \|v\|_{L_2}=\rho\} \subset {\cal K}^c$ and by homogeneity, $(\rho {\cal B})^c \subset {\cal K}^c$, as required.

\endproof

Once Theorem \ref{thm:main-1} is established, one may apply it to random vectors that satisfy $(1)$ and $(2)$ --- for example, the standard gaussian vector or the vector distributed uniformly on $S^{d-1}$. It follows that for any $\eta \leq c_0$ and given more than $c_1d\eta^{-2} \log (2/\eta)$ random points, the random set ${\cal K}$ is a $c_2\eta$-approximation of ${\cal B}$ for an absolute constant $c_2$. In particular, Theorem \ref{thm:zig-zag} follows from Theorem \ref{thm:main-1}.

\vskip0.4cm
Clearly, since a general random vector $X$ need not satisfy $(1)$ and $(2)$, the proof of Theorem \ref{thm:main-intro} requires an additional step. To that end one may invoke the Berry-Esseen Theorem (see, e.g., \cite{MR2722836}) to `smooth' $X$ and construct a random vector $Z$ that does satisfy $(1)$ and $(2)$.

\begin{Theorem} \label{thm:Berry-Esseen}
Let $W$ be a mean-zero random variable and let $W_1,...,W_m$ be independent copies of $W$. If
$$
Y=\frac{1}{\sqrt{m}\|W\|_{L_2}} \sum_{i=1}^m W_i,
$$
then
$$
\sup_{t \in \R} \left|Pr(Y > t) - Pr(g>t) \right| \leq \psi(m),
$$
where $\psi(m)=C (\|W\|_{L_3}^3/\|W\|_{L_2}^3)m^{-1/2}$. In particular, if $\|W\|_{L_3} \leq L \|W\|_{L_2}$ then $\psi(m)=c(L)/\sqrt{m}$.
\end{Theorem}

\begin{Remark}
There are other versions of the Berry-Esseen Theorem with different conditions on $W$. For example, one may obtain nontrivial estimates on $\psi(m)$ as soon as $\|W\|_{L_q} \leq L \|W\|_{L_2}$ for some $q>2$, although if $2<q<3$ then $\psi(m)$ tends to $0$ at a slower (polynomial) rate than $1/\sqrt{m}$ (see \cite{MR2630040}). Alternatively, if $Y \in L_{\psi_\alpha}$, one has better estimates on $\psi(m)$ (see, e.g., \cite{MR1353441}).
\end{Remark}

For an integer $m \leq N$, set
\begin{equation} \label{eq:Z}
Z=\frac{1}{\sqrt{m}} \sum_{i=1}^m X_i,
\end{equation}
and thus one has access to $n=N/m$ independent copies of $Z$. Clearly, $Z$ is centred and has the same covariance structure as $X$. Also, for any $v \in {\cal S}$,
$$
\sup_{t \in \R} \left|Pr(|\inr{Z,v}| \leq t) - Pr(|g| \leq t) \right| \leq 2\psi(m).
$$
Therefore, if we set $\alpha$ to be the median of $|g|$, then for every $v \in {\cal S}$,
\begin{equation} \label{eq:(1)-general}
\left|Pr(|\inr{Z,v}| \leq \alpha)-\frac{1}{2} \right| \leq 2\psi(m).
\end{equation}
Moreover, if $\eps \leq \alpha/2$, there is an absolute constant $c$ for which
\begin{align*}
& Pr(|\inr{Z,v}| \leq \alpha-\eps) \leq Pr(|g| \leq \alpha-\eps) + 2\psi(m) = Pr(|g| \leq \alpha) - c\eps +2\psi(m)
\\
\leq & Pr(|\inr{Z,v}| \leq \alpha) -c\eps + 4\psi(m);
\end{align*}
Hence, if $\eps \geq 8\psi(m)/c$, it follows that
$$
Pr(|\inr{Z,v}| \in [\alpha-\eps,\alpha]) \geq c^\prime \eps
$$
for an absolute constant $c^\prime$.

Thus, Condition $(2)$ holds for $\eps_0 =8\psi(m)/c$ and $\eps$ that satisfies $\eps_0 \leq \eps \leq 1/8$; clearly, $\eps_0$ can be made arbitrarily small by taking a large enough $m$.

\vskip0.4cm
\noindent{\bf Proof of Theorem \ref{thm:main-intro}.}
Given the wanted accuracy parameter $\eta$, let $m$ for which $8\psi(m)/c \leq \eta \leq 1/8$. By Theorem \ref{thm:Berry-Esseen}, if $q \geq 3$ and  $\sup_{v \in {\cal S}} \|\inr{X,v}\|_{L_q} \leq L$ then one may take $m=c(L)/\eta^2$, whereas by \cite{MR2630040}, if $2<q<3$ one may take $m=c(L){\rm poly}(1/\eta)$.

Define $Z$ as in \eqref{eq:Z} and take $Z_1,...,Z_n$ to be $n$ independent copies of $Z$ for $n \geq c_1 \eta^{-2}\log(2/\eta)d$. Set $\alpha$ to be the median of $|g|$; by Theorem \ref{thm:main-1}, with probability at least $1-2\exp(-c_2\eta^2 n)$, the random set ${\cal K}$ satisfies
$$
{\cal B} \subset {\cal K} \subset (1+c_3\eta){\cal B},
$$
as required.
\endproof

%%%%%%%%%%%%%%%%%%%%%%%%%%%%%%%%%%%%%%%%%%%%%%%%%%%%%%%%%%%%%%%5
\subsubsection{Isomorphic approximation}

If one is interested in an isomorphic approximation, i.e., that $c{\cal B} \subset {\cal K} \subset C{\cal B}$ for constants $c$ and $C$ that need not be close to $1$, the assumption required in Theorem \ref{thm:main-intro} can be relaxed from norm equivalence to a small-ball condition: that there are $0<\lambda,\delta<1$ such that for every $v \in \R^d$,
\begin{equation} \label{eq:small-ball-cond}
Pr(|\inr{X,v}| \geq \lambda \|v\|_{L_2}) \geq \delta.
\end{equation}
By a similar argument to the one used in the proof of Theorem \ref{thm:main-intro} it follows that for
$$
N \gtrsim \max\left\{\frac{d}{\delta}\log(1/\delta \lambda), \frac{d}{\lambda^2}\right\},
$$
and setting
$$
{\cal K} = \{v \in \R^d : |\inr{X_i,v}| \leq \lambda/2 \ {\rm for \ at \ least \ } (1-\delta/4) N \ {\rm indices} \ i \},
$$
with probability at least $1-2\exp(-c\delta N)$,
$$
c^\prime \lambda \sqrt{\delta} {\cal B} \subset {\cal K} \subset {\cal B}.
$$
The inclusion ${\cal K} \subset {\cal B}$ stems from the small-ball condition: for every $v \in {\cal S}$, with probability at least $1-2\exp(-cN)$, at least $\delta N/2$ of the values $|\inr{X_i,v}|$ are likely to be larger than $\lambda$. The reason behind the other inclusion, that $c^\prime \lambda \sqrt{\delta} {\cal B} \subset {\cal K}$, is that $Pr(|\inr{X,v}| \geq t\|v\|_{L_2}) \leq 1/t^2$; therefore, with probability at least $1-2\exp(-cN)$, most of the values $|\inr{X_i,v}|$ cannot be `too large'. The high probability with which both properties hold allows one to control a fine enough net in the sphere, and the oscillation term is handled in a similar way to the proof of Theorem \ref{thm:main-1}. We omit the straightforward details.

\subsection{Approximation using ellipsoids} \label{sec:ellipsoids}
This section is devoted to the proof of Theorem \ref{thm:ellipsoids}. Let $m$ to be specified in what follows, set $n=N/m$ and let $I_1,...,I_n$ be the natural decomposition of $\{1,...,N\}$ to coordinate blocks of cardinality $m$.

For $1 \leq j \leq n$ and $v \in \R^d$ set
$$
Z_j(v)=\frac{1}{m}\sum_{i \in I_j} \inr{X_i,v}^2
$$
and recall that
$$
{\cal D}_{\eta} = \{v \in \R^d : |\{j : Z_j(v) \leq 1+\eta \}| \geq 0.9n\}.
$$
Our aim is to show that if $m$ and $n$ are chosen properly, then with high probability,
$$
{\cal B} \subset {\cal D}_\eta \subset (1+c\eta){\cal B}
$$
for a suitable absolute constant $c$.

It is important to stress that the natural candidate for approximating ${\cal B}$, the empirical $L_2$ ball
$$
\left\{ v \in \R^d : \frac{1}{N}\sum_{i=1}^N \inr{X_i,v}^2 \leq 1\right\},
$$
can be very different from ${\cal B}$ when $X$ is heavy-tailed; this will be illustrated in Section \ref{sec:example}.

\vskip0.4cm

Again, we follow the general path outlined in Section \ref{sec:small-ball}. The property ${\cal P}$ is given by invoking  Assumption \ref{ass:ellipsoids}---that if $m=m_0(\eta)$ then for every $v \in {\cal S}$
$$
Pr\left( \left|\frac{1}{m} \sum_{i=1}^m \inr{X_i,v}^2 - 1  \right| \geq \frac{\eta}{10}\right) \leq 0.01.
$$

\begin{Theorem} \label{thm:L-2-uniform}
There are absolute constants $c_1$ and $c_2$ for which the following holds. If
$$
n \geq c_1\max\{d \log(2m_0(\eta)/\eta),m_0(\eta)\},
$$
then with probability at least $1-2\exp(-c_2n/m_0(\eta))$, for every $v \in \R^d$
\begin{equation} \label{eq:global}
|\{ j : Z_j(v) \in [(1-\eta)\E Z(v),(1+\eta)\E Z(v)] \} | \geq 0.96n.
\end{equation}
In particular, if $m_0(\eta) \leq C\eta^{-k}$ then $n \geq c_1(k+1)d\log(2C/\eta)$ suffices.
\end{Theorem}

\begin{Corollary}
It is straightforward to verify that under an $L_4-L_2$ norm equivalence with constant $L$ one has that $m_0(\eta) \leq c(L)/\eta^2$. Therefore, the required sample size is $N=m_0 n$ for
$$
m_0 \leq c_1(L)\eta^{-2} \ \ \ {\rm and} \ \ \ n = c^\prime(L) \max\{d \log (L/\eta),\eta^{-2}\}
$$
which is a better estimate than in Theorem \ref{thm:main-intro} as long as $\eta \gtrsim 1/(d \log d)^{1/2}$.
\end{Corollary}

\proof Since the claim is homogeneous in $v$ it suffices to show that it holds for $v \in {\cal S}$. By a binomial estimate, there is an absolute constant $c_0$ such that each $v \in \R^d$ satisfies
\begin{equation} \label{eq:single}
|\{j : Z_j(v) \in [(1-\eta/10)\E Z , (1+\eta/10) \E Z]\}| \geq 0.98n
\end{equation}
with probability at least $1-2\exp(-c_0n)$.

Let $V \subset {\cal S}$ be of cardinality at most $\exp(c_0n/2)$. Invoking the probability estimate with which \eqref{eq:single} holds, there is an event ${\cal A}_1$ of probability at least $1-2\exp(-c_0n/2)$ such that \eqref{eq:single} holds for every $v \in V$. As expected, our choice of $V$ is a maximal $r$-separated subset of ${\cal S}$ with respect to the $L_2$ norm; and by a volumetric estimate, $r \leq 5\exp(-c_1n/d)$ for an absolute constant $c_1$.

To prove the wanted uniform estimate, for $u \in {\cal S}$ let $\pi u \in V$ be the nearest element to $u$ with respect to the $L_2$ norm. Set
$$
W = \sup_{u \in {\cal S}} |\{i : |\inr{X_i,u-\pi u}| \geq \eta/10\}|,
$$
and the aim is to show that with high probability, $W \leq 0.02n$.

Just as in the proof of Theorem \ref{thm:main-1}, let us first estimate $\E W$. By symmetrization and contraction, followed by the estimate on $r$ and Lemma \ref{lemma:Bernoulli},
\begin{align*}
\E W \leq & \frac{10}{\eta} \E \sup_{u \in {\cal S}} \left|\sum_{i=1}^N \left(|\inr{X_i,u-\pi u}| - \E |\inr{X_i,u-\pi u}|\right)\right| + \frac{10}{\eta}\sup_{u \in {\cal S}} |\inr{X,u-\pi u}|
\\
\leq & \frac{20r}{\eta} \left(\E \sup_{u \in {\cal B}} \left|\sum_{i=1}^N \eps_i \inr{X_i,u} \right| + N \right) \leq c_2\frac{rN}{\eta}\left(\sqrt{dN} + N \right) \leq 0.01n,
\end{align*}
provided that $n \geq c_3d \log(m_0(\eta)/\eta)$. Therefore, by the bounded differences inequality, $W \leq 0.02n$ with probability at least $1-2\exp(-c_4n^2/N)=1-2\exp(-c_4n/m)$ for a suitable absolute constant $c_4$.

Combining the two estimates, there is an event with probability at least $1-2\exp(-c_5n/m)$ on which:
\begin{description}
\item{$\bullet$} For every $v \in V$, $Z_j(v) \in [1-\eta/10,1+\eta/10]$ for at least $0.98n$ indices $j$.
\item{$\bullet$} For every $u \in {\cal S}$, $|\inr{X_i,u-\pi u} | \geq \eta/10$ for at most $0.02n$ indices $i$; in particular, for every $u$ there could be at most $0.02n$ of the coordinate blocks $I_j$ that are `corrupted' by such a large value of $|\inr{X_i,u-\pi u}| \geq \eta/10$. On all the other blocks, $\max_{i \in I_j} |\inr{X_i,u-\pi,u}| \leq \eta/10$.
\end{description}
Therefore, by the triangle inequality, for every $u \in {\cal S}$ there are at least $0.96n$ indices $j$ for which $Z_j(u) \in [1-\eta,1+\eta]$, as required.
\endproof

\noindent{\bf Proof of Theorem \ref{thm:ellipsoids}.} Consider the event from Theorem \ref{thm:L-2-uniform}. If $u \in {\cal S}$ then $Z_j(u) \leq 1+\eta$ for more than $0.9n$ coordinate blocks, implying that $u \in {\cal D}_\eta$. And, since ${\cal D}_\eta$ is star-shaped around $0$, it is evident that ${\cal B} \subset {\cal D}_\eta$.

At the same time, if $\|u\|_{L_2}= \rho$ then $Z_j(u) \geq (1-\eta) \rho^2 > 1+\eta$ provided that $\rho \geq 1+c\eta$. Therefore, $(1+c\eta){\cal S} \subset ({\cal D}_\eta)^c$ and in particular, using the star-shape property again, ${\cal D}_\eta \subset (1+c\eta){\cal B}$.
\endproof

\subsection{Limitations of approximating using the empirical ellipsoid } \label{sec:example}
Let us show that selecting ${\cal K}=\{v \in \R^d : N^{-1}\sum_{i=1}^N \inr{X_i,v}^2 \leq 1\}$ as an approximation of ${\cal B}$ is a poor choice when $X$ is heavy-tailed.
To that end we construct a collection of random vectors that satisfy an $L_4-L_2$ norm equivalence and for which ${\cal B}$ is equivalent to $B_2^d$. At the same time, with a non-trivial probability there is $v \in S^{d-1}$ for which $N^{-1}\sum_{i=1}^N \inr{X_i,v}^2 \gg 1$. More accurately, for each $u \gtrsim 1/\sqrt{d}$ we construct a centred random vector $X_u$ that satisfies:
\begin{description}
\item{$(a)$} For every $v \in S^{d-1}$, $1 \leq \|\inr{X_u,v}\|_{L_2} \leq 2$;
\item{$(b)$} $\sup_{v \in S^{d-1}} \|\inr{X,v}\|_{L_4} \leq L$ for an absolute constant $L$; and
\item{$(c)$} $Pr(\|X_u\|_2^2 \geq ud) \geq 1/2u^2d$.
\end{description}

Let $\Gamma=N^{-1/2}\sum_{i=1}^N \inr{X_i,\cdot}e_i$ and observe that
\begin{equation*}
\sup_{v \in S^{d-1}} \frac{1}{N}\sum_{i=1}^N \inr{X_i,v}^2 = \|\Gamma\|^2_{2 \to 2} = \|\Gamma^*\|_{2 \to 2}^2 \geq \max_{1 \leq i \leq N} \|\Gamma^* e_i\|_2^2
\geq \frac{1}{N}\max_{1 \leq i \leq N} \|X_i\|_2^2.
\end{equation*}

\begin{Lemma} \label{eq:lemma-X-u}
Let $0<\delta<1/4$ and set $X_u$ as above for $u=(N/4d \delta)^{1/2}$. Then with probability at least $\delta$,
$$
\frac{1}{N} \max_{1 \leq i \leq N} \|X_i\|_2^2 \geq  \sqrt{\frac{d}{4\delta N}}.
$$
\end{Lemma}
In particular, with probability at least $\delta$, $B_2^d \not \subset C{\cal K}$ unless $C \geq (d/4N\delta)^{1/4}$, making even an isomorphic approximation impossible if one would like it to hold with probability $1-\delta$ for a small $\delta$ (corresponding to a large $u$), particularly taking into account that we would like $N$ to scale linearly in $d$.

\vskip0.4cm

\proof Recall that $Pr(\|X_u\|_2^2 \geq ud) \geq 1/2u^2d=2\delta/N \equiv \rho$. Therefore, given $N$ independent copies of $X_u$ denoted by $Y_1,...,Y_N$,
$$
Pr({\rm there \ exists \ } 1 \leq i \leq N, \ \|Y_i\|^2 \geq ud) \geq N \rho (1-\rho)^{N-1} = 2\delta (1-2\delta/N)^N \geq \delta.
$$
On that event,
$$
\frac{1}{N} \max_{1 \leq i \leq N} \|Y_i\|_2^2 \geq \frac{ud}{N} = \left(\frac{d}{4N\delta}\right)^{1/2},
$$
as claimed.
\endproof

All that is left now is to construct the random vectors $X_u$. To that end, let $\eta_1,...,\eta_d$ be independent $\{0,1\}$-valued random variables with mean $1/u^2 d^2$ and set $\eps_1,...,\eps_d$ to be independent, symmetric $\{-1,1\}$-valued random variables that are independent of $\eta_1,...,\eta_d$. Let $z_i=\eps_i\max\{\eta_i R,1\}$ where $R = \sqrt{ud}$, and set $X_u=(z_1,...,z_d)$.

Clearly, $\E z_i =0$ and
$$
\E z_i^2 = \frac{R^2}{u^2 d^2} + \left(1-\frac{1}{u^2d^2}\right);
$$
hence, $1 \leq \|z_i\|_{L_2} \leq 2$ if $u \geq 1/d$ as was assumed. Moreover,
$$
\E z_i^4 \leq \frac{R^4}{u^2 d^2} + \left(1-\frac{1}{u^2d^2}\right) \leq 2.
$$
Now, for $v \in \R^d$ we have that
$\E\inr{X_u,v}^2 = \sum_{i=1}^d v_i^2 \E z_i^2$ and $(a)$ follows from the estimate on $\E z_i^2$. As for $(b)$, it is straightforward to verify that since $\E z_i^4 \leq 2$, $\|\sum_{i=1}^d v_i z_i\|_{L_4} \leq L\|v\|_2$ for an absolute constant $L$. Finally, to prove $(c)$, consider $u \gtrsim 1/\sqrt{d}$ and observe that $\|X_u\|_2^2 = \sum_{i=1}^d z_i^2$. Note that with probability at least $d \cdot (1/u^2d^2) \cdot (1-1/u^2d^2)^{d-1} \geq 1/2u^2 d$, there is at least one index $i$ for which $z_i^2 \geq R^2 = ud$; hence, on that event, $\|X_u\|_2^2 \geq ud$, as required.
\endproof

\subsection{Improving Theorem \ref{thm:main-intro}} \label{sec:improve}
Let us sketch an alternative proof of Theorem \ref{thm:main-intro}. On the one hand, it leads to a better estimate on the required sample size; on the other, it is based on a special property of slabs. The components of the proof are well understood so we will only sketch the argument.

\vskip0.4cm

In what follows we consider $Z_1,...,Z_n$ that are distributed as $m^{-1/2} \sum_{i=1}^m X_i$ and satisfy \eqref{eq:(1)-general}; specifically we assume that $m$ is large enough to ensure that for $v \in {\cal S}$, 
\begin{equation} \label{eq:emp-proof-2}
\left|Pr(|\inr{Z,v}| \leq \alpha)-\frac{1}{2} \right| \leq \frac{\eta}{2}
\end{equation}
where $\alpha$ is the median of $|g|$.

Here, the approximating body will be
$$
{\cal K} = \left\{v \in \R^d : |\inr{Z_j,v}| \leq \alpha \ {\rm for \ at \ least \ } \left(\frac{1}{2}-\eta\right)n \ {\rm indices} \ j\right\}.
$$

To show that indeed ${\cal K}$ is an $\eta$-approximation of ${\cal B}$, let us estimate the supremum of the empirical process
\begin{equation} \label{eq:empirical}
W=\sup_{v \in {\cal S}} \left|\frac{1}{n} \sum_{j=1}^n \IND_{\{|\inr{Z_j,v}| \leq \alpha\}} - Pr ( |\inr{Z,v}| \leq \alpha ) \right|.
\end{equation}
This is an empirical process indexed by a collection ${\cal U}$ of subsets of $\R^d$---the slabs $\{x \in \R^d : |\inr{x,v}| \leq \alpha\}$. It is standard to verify that the \emph{VC dimension} of ${\cal U}$ is at most $cd$:  each set is generated by the intersection of two halfspaces, and the VC dimension of the collection of halfspaces in $\R^d$ is at most $c^\prime d$ (see, for example, \cite{vaWe96} for more information on VC classes).

By Talagrand's concentration inequality for empirical processes indexed by a class of bounded functions (\cite{MR1258865}, see also \cite{BoLuMa13}), it follows that with probability at least $1-\exp(-t)$,
$$
W \leq c_1\left(\E W + \sqrt{\frac{t}{n}} + \frac{t}{n}\right).
$$
And, by a standard argument\footnote{The proof is based on symmetrization, the fact that a Bernoulli process is subgaussian with respect to the $\ell_2$ metric, a Dudley entropy integral bound  and well-known estimates on the covering numbers of VC-classes.},
\begin{equation*}
\E W \leq c_2\sqrt{\frac{d}{n}}.
\end{equation*}
Thus, with probability at least $1-\exp(-c_3\eta^2n)$, $W \leq \eta/2$
provided that $n \gtrsim d/\eta^2$.

Therefore, on that event
\begin{equation} \label{eq:emp-proof-1}
\sup_{v \in {\cal S}} \left| \left| \left\{j : |\inr{Z_j,v}| \leq \alpha\right\} \right| - n Pr(|\inr{Z,v}| \leq \alpha) \right| \leq \frac{n\eta}{2}.
\end{equation}

Combining \eqref{eq:emp-proof-1} and \eqref{eq:emp-proof-2} it follows that with probability at least $1-2\exp(-c\eta^2n)$, for every $v \in {\cal S}$,
\begin{equation} \label{eq:emp-proof-3}
\left| \left\{j : |\inr{Z_j,v}| \leq \alpha\right\} \right| \geq n\left(\frac{1}{2}-\eta\right).
\end{equation}
In particular we have that ${\cal S} \subset {\cal K}$, and since ${\cal K}$ is star-shaped around $0$ then also ${\cal B} \subset {\cal K}$.

A similar estimate to \eqref{eq:emp-proof-3} leads to the fact that $(1+\eta){\cal S} \subset {\cal K}^c$ and completes the proof.
\endproof

\vskip0.4cm
The feature that makes this proof simple is that the class of indicators one is interested in happens to be a VC class. In general, there is no reason to expect such a happy coincidence when choosing a property ${\cal P}$, and controlling the resulting empirical process can be a nontrivial problem. In contrast, the method presented here allows one by bypass this difficulty for rather general choices of ${\cal P}$ and at a price of a slightly suboptimal dependency on $\eta$.

\bibliographystyle{plain}
\bibliography{zigzag}

\begin{thebibliography}{10}

\bibitem{MR2601042}
Rados{\l}aw Adamczak, Alexander~E. Litvak, Alain Pajor, and Nicole
  Tomczak-Jaegermann.
\newblock Quantitative estimates of the convergence of the empirical covariance
  matrix in log-concave ensembles.
\newblock {\em J. Amer. Math. Soc.}, 23(2):535--561, 2010.

\bibitem{MR2204286}
S.~Artstein-Avidan, O.~Friedland, and V.~Milman.
\newblock Geometric applications of {C}hernoff-type estimates and a zigzag
  approximation for balls.
\newblock {\em Proc. Amer. Math. Soc.}, 134(6):1735--1742, 2006.

\bibitem{BoLuMa13}
S.~Boucheron, G.~Lugosi, and P.~Massart.
\newblock {\em Concentration inequalities: A Nonasymptotic Theory of
  Independence}.
\newblock Oxford University Press, 2013.

\bibitem{MR1761898}
Gerald H.~L. Cheang and Andrew~R. Barron.
\newblock A better approximation for balls.
\newblock {\em J. Approx. Theory}, 104(2):183--203, 2000.

\bibitem{MR2722836}
Rick Durrett.
\newblock {\em Probability: theory and examples}, volume~31 of {\em Cambridge
  Series in Statistical and Probabilistic Mathematics}.
\newblock Cambridge University Press, Cambridge, fourth edition, 2010.

\bibitem{MR757767}
Evarist Gin\'e and Joel Zinn.
\newblock Some limit theorems for empirical processes.
\newblock {\em Ann. Probab.}, 12(4):929--998, 1984.
\newblock With discussion.

\bibitem{MR670396}
Peter~M. Gruber and Petar Kenderov.
\newblock Approximation of convex bodies by polytopes.
\newblock {\em Rend. Circ. Mat. Palermo (2)}, 31(2):195--225, 1982.

\bibitem{LeTa91}
M.~Ledoux and M.~Talagrand.
\newblock {\em Probability in Banach Space}.
\newblock Springer-Verlag, New York, 1991.

\bibitem{MenACM}
S.~Mendelson.
\newblock Learning without concentration.
\newblock {\em Journal of the ACM}, 62:21, 2015.

\bibitem{MR3191978}
Shahar Mendelson and Grigoris Paouris.
\newblock On the singular values of random matrices.
\newblock {\em J. Eur. Math. Soc. (JEMS)}, 16(4):823--834, 2014.

\bibitem{MR2630040}
Elchanan Mossel, Ryan O'Donnell, and Krzysztof Oleszkiewicz.
\newblock Noise stability of functions with low influences: invariance and
  optimality.
\newblock {\em Ann. of Math. (2)}, 171(1):295--341, 2010.

\bibitem{MR1353441}
Valentin~V. Petrov.
\newblock {\em Limit theorems of probability theory}, volume~4 of {\em Oxford
  Studies in Probability}.
\newblock The Clarendon Press, Oxford University Press, New York, 1995.
\newblock Sequences of independent random variables, Oxford Science
  Publications.

\bibitem{MR608101}
Rolf Schneider and John~Andr\'e Wieacker.
\newblock Approximation of convex bodies by polytopes.
\newblock {\em Bull. London Math. Soc.}, 13(2):149--156, 1981.

\bibitem{MR3127875}
Nikhil Srivastava and Roman Vershynin.
\newblock Covariance estimation for distributions with {$2+\varepsilon$}
  moments.
\newblock {\em Ann. Probab.}, 41(5):3081--3111, 2013.

\bibitem{MR1258865}
M.~Talagrand.
\newblock Sharper bounds for {G}aussian and empirical processes.
\newblock {\em Ann. Probab.}, 22(1):28--76, 1994.

\bibitem{Tikh}
Konstantin Tikhomirov.
\newblock Sample covariance matrices of heavy-tailed distributions.
\newblock {\em IMRN}, to appear.

\bibitem{vaWe96}
A.W. {van der Vaart} and J.A. Wellner.
\newblock {\em Weak convergence and empirical processes}.
\newblock Springer-Verlag, New York, 1996.

\end{thebibliography}

\end{document}